\documentclass{article}
\usepackage[utf8]{inputenc}
\usepackage{authblk}
\usepackage{tabularx}
\usepackage{graphicx}
\usepackage{bbold}
\usepackage{amstext}
\usepackage{amssymb}
\usepackage{arydshln}
\usepackage{array}
\usepackage{hyperref}

\title{Evaluation of Deep Species Distribution Models using Environment and Co-occurrences}
\author[1]{Benjamin Deneu}
\author[2]{Maximilien Servajean}
\author[3]{Christophe Botella}
\author[1]{Alexis Joly}

\affil[1]{Inria, LIRMM, Montpellier, France, \href{mailto:benjamin.deneu@inria.fr}{benjamin.deneu@inria.fr}, \href{mailto:alexis.joly@inria.fr}{alexis.joly@inria.fr}}

\affil[2]{AMIS, Université Paul Valéry Montpellier, LIRMM UMR 5506, CNRS, \href{mailto:maximilien.servajean@lirmm.fr}{maximilien.servajean@lirmm.fr}}

\affil[3]{INRA, Inria, AMAP, LIRMM, Montpellier, France, \href{mailto:christophe.botella@inria.fr}{christophe.botella@inria.fr}}

\date{}

\begin{document}

\maketitle

\begin{abstract}

This paper presents an evaluation of several approaches of plants species distribution modeling based on spatial, environmental and co-occurrences data using machine learning methods. In particular, we re-evaluate the environmental convolutional neural network model that obtained the best performance of the GeoLifeCLEF 2018 challenge but on a revised dataset that fixes some of the issues of the previous one. We also go deeper in the analysis of co-occurrences information by evaluating a new model that jointly takes environmental variables and co-occurrences as inputs of an end-to-end network. Results show that the environmental models are the best performing methods and that there is a significant amount of complementary information between co-occurrences and environment. Indeed, the model learned on both inputs allows a significant performance gain compared to the environmental model alone.

\end{abstract}

\section{Introduction}
Automatically predicting the list of species that are the most likely to be observed at a given location is useful for many scenarios in biodiversity informatics. First of all, it could improve species identification processes and tools by reducing the list of candidate species that are observable at a given location (be they automated, semi-automated or based on classical field guides or flora). More generally, it could facilitate biodiversity inventories through the development of location-based recommendation services (typically on mobile phones) as well as the involvement of non-expert nature observers. Last but not least, it might serve educational purposes thanks to biodiversity discovery applications providing innovative features such as contextualized educational pathways.\\
\indent This problem is known as Species Distribution Modeling (SDM) in ecology. SDM have become increasingly important in the last few decades for the study of biodiversity, macro ecology, community ecology and the ecology of conservation. An accurate knowledge of the spatial distribution of species is actually of crucial importance for many concrete scenarios including landscape management, preservation of rare and/or endangered species, surveillance of alien invasive species,  measurement of human impact or climate change on species, etc. Concretely, the goal of SDM is to infer the spatial distribution of a given species, and they are often based on a set of geo-localized occurrences of that species (collected by naturalists, field ecologists, nature observers, citizen sciences project, etc.). However, it is usually not reliable to learn that distribution directly from the spatial positions of the input occurrences. The two major problems are the limited number of occurrences and the bias of the sampling effort compared to the real underlying distribution. In a real-world dataset, the raw spatial distribution of the occurrences is actually highly influenced by the accessibility of the sites and the preferences and habits of the observers. Another difficulty is that an occurrence means a punctual presence of the species, while no occurrences doesn't mean the species is absent, which makes us very uncertain about regions without observed specimens.\\
\indent For all these reasons, SDM is usually achieved through \textit{environmental niche modeling} approaches, \textit{i.e.} by predicting the distribution in the geographic space on the basis of a representation in the environmental space \cite{Guisan2000,Guisan2005,Araujo2006,Elith2006,Franklin2010,Peterson2011,Ferrarini2019}. This environmental space is in most cases represented by climate data (such as temperature, and precipitation), but also by other variables such as soil type, land cover, distance to water, etc. Then, the objective is to learn a function that takes the environmental feature vector of a given location as input and outputs an estimate of the abundance of the species. The main underlying hypothesis is that the abundance function is related to the \textit{fundamental ecological niche} of the species. That means that in theory, a given species is likely to live in a privileged ecological niche, characterized by an hypervolume in the environmental space. However, this volume can have a very irregular shape and, in addition, many phenomena can actually affect the distribution of the species relative to its so called \textit{abiotic} preferences. The real distribution of the species is called \textit{realized ecological niche} it can differ from the \textit{fundamental ecological niche} by environmental perturbations, geographical constraints, or interactions with other living organisms (including humans) that might have encourage specimens of that species to live in a different environment. As a consequence, the \textit{realized ecological niche} of a species can be much more diverse and complex than its hypothetical fundamental niche.\\
\indent Very recently, SDM based on deep neural networks have started to appear \cite{mtapbook2018botella}. These first experiments showed that they can have a good predictive power, potentially better than the models used conventionally in ecology. Actually, deep neural networks are able to learn complex nonlinear transformations in a wide variety of domains. In addition, they make it possible to learn an area of environmental representation common to a large number of species, which stabilizes predictions from one species to another and improves them globally \cite{Pollock2014}. Finally, spatial patterns in environmental variables often contain useful information for species distribution but are generally not considered in conventional models. Conversely, convolutional neural networks effectively use this information and improve prediction performance.\\
\indent In this paper, we report an evaluation study of four main kinds of plants SDM:
\begin{enumerate}
    \item A convolutional neural network aimed at learning the ecological preferences of species thanks to environmental image patches provided as inputs (temperature, soil type, etc.).
    \item Two purely spatial models, one based on a random forest fitted on the spatial coordinates of the occurrences of each species. The other named closest-locations classifier is close to a nearest neighbours classifier.
    \item A species co-occurrence model aiming at predicting the likelihood of presence of a given species thanks to the knowledge of the presence of other species. 
    \item We finally introduce a new neural network that jointly learn on environment and co-occurrences and compare it to the two separated models to study the joint information between the environment and co-occurrences. And in parallel to it, a model consisting of a late merging of the outputs of the co-occurrences model and the environmental CNN.
\end{enumerate}
This paper is an extended and revised version of the working note that we wrote beforehand in the context of our participation to the GeoLifeCLEF 2018 challenge \cite{deneugeolifeclef2018}. It improves it two main ways. First we re-evaluate the environmental convolutional neural network models that obtained the best performance during the GeoLifeCLEF 2018 challenge but on a revised dataset that fixes some of the issues of the previous one (that were discovered after the end of the challenge). We also go deeper in the analysis of co-occurrences information by evaluating a new model that jointly takes environmental variables and co-occurrences as inputs of an end-to-end network. Section \ref{data} gives an overview of the data and evaluation methodology. Section \ref{models} and \ref{fusion model} provide the detailed description of the evaluated models. Section \ref{expes} presents the results of the experiments and their analysis.\\

\section{Data and Evaluation Methodology}
\label{data}
A detailed description of the protocol used to build the GeoLifeCLEF2018 dataset is provided in \cite{geolifeclef2018,lifeclef2018}. In a nutshell, the dataset was built from occurrence data of the Global Biodiversity Information Facility (GBIF), the world’s largest open data infrastructure in this domain, funded by governments. It is composed of $291,392$ occurrences of $N=3,336$ plant species observed on the French territory between 1835 and 2017. This dataset was split in 3/4 for training and 1/4 for testing with the constraints that: (i) for each species in the test set, there is at least one observation of it in the train set and (ii), an observation of a species in the test set is distant of more than 100 meters from all observations of this species in the train set.\\
\indent Concerning the environmental images used to learn CNN models, those given during the 2018 session of the challenge were found to be erroneous and enabled the models to actually overfit a part of the test set (mostly the occurrences that overlapped the sea). To resolve this problem, new environmental images of each occurrence were extracted using the patch extractor from GeoLifeCLEF2019 \footnote{\url{https://github.com/maximiliense/GLC19}}. The environmental data is composed of 33 environmental rasters. Each raster encodes an environmental variable on the French territory. They were constructed from various open datasets including Chelsea Climate, ESDB soil pedology data, Corine Land Cover 2012 soil occupation data, CGIAR-CSI evapotranspiration data, USGS Elevation data (Data available from the U.S. Geological Survey.) and BD Carthage hydrologic data. To construct the input tensor we extract for each occurrence a matrix of $64\times64$ pixels from each raster centered of the location of the occurrence. Most of the environmental variables are continuous variables such as the average temperature, the altitude or the distance to water. Thus, the corresponding $64\times64$ pixel matrices can be processed as classical image channels provided as input of the CNN. Some of the variables are rather of ordinal type (such as ESDB v2). However, they still can be considered as additional channels of the CNN in the sense that the order of the pixel values remains meaningful. This is not true, however, for categorical variables such as the Corine Land Cover variable. This variable take up to $45$ different categorical values but the order of these values does not have any meaning. Consequently, this patch is unstacked into 45 different binary patches. We finally obtain a tensor of size $64\times 64\times(32+45=77)$ for each occurrence.\\
\indent In the following, we usually denote as $x \in X$ a particular occurrence, each $x$ being associated to a spatial position $p(x)$ in the spatial domain $D$, a species label $y(x)$ and an environmental tensor $\textbf{g}(x)$ of size $64\times64\times77$. We denote as $P$ the set of all spatial positions $p$ covered by $X$. It is important to note that a given spatial position $p \in P$ usually corresponds to several occurrences $x_j \in X, p(x_j)=p$ observed at that location (18 000 spatial locations over a total of 60 000, because of quantized GPS coordinates or Names-to-GPS transforms). In the training set, up to several hundreds of occurrences can be located at the same place (be they of the same species or not). The occurrences in the test set might also occur at identical locations but, by construction, the occurrence of a given species does never occur at a location closer than 100 meters from the occurrences of the same species in the training set.\\
\indent The used evaluation metric is the Mean Reciprocal Rank (MRR). The MRR is a statistic measure for evaluating any process that produces a list of possible responses ordered by probability of correctness. It is well adapted to assess the scenario targeted by the GeoLifeCLEF challenge, \textit{i.e.} providing a short-list of species that are the most likely to be observed at a given location to users of field applications. The reciprocal rank of a query response is the multiplicative inverse of the rank of the first correct answer. The MRR is the average of the reciprocal ranks for the whole test set:
\begin{equation}
    MRR = \frac{1}{Q} \sum_{q=1}^{Q}\frac{1}{\text{rank}_q}
\end{equation}
where $Q$ is the total number of query occurrences $x_q$ in the test set and $rank_q$ is the rank of the correct species $y(x_q)$ in the ranked list of species predicted by the evaluated method for $x_q$. \\

\section{Evaluated SDM Models in GeoLifeCLEF2018}
\label{models}

\subsection{Convolutional Neural Network}
\label{CNN}
It has been previously shown in \cite{mtapbook2018botella} that Convolutional Neural Networks (CNN) may reach better predictive performance than classical models used in ecology. Our approach builds upon this idea but differs from the one of Botella \textit{et al} in two important points:
\begin{itemize}
    \item \textbf{Softmax loss}: whereas the CNN of Botella \textit{et al.} \cite{mtapbook2018botella} was aimed at predicting species abundances thanks to a Poisson regression on the learned environmental features, our model rather attempts to predict the most likely species to be observed according to the learned environmental features. In practice, this is simply done by using a softmax layer and a categorical loss instead of the Poisson loss layer used in \cite{mtapbook2018botella}.
    \item \textbf{Model architecture}: we also used a different architecture of the convolutional layers compared to the one of Botella \textit{et al.} and the one submitted during our participation in the challenge \cite{deneugeolifeclef2018}. We used the inception v3 architecture \cite{DBLP:journals/corr/SzegedyVISW15} but with the three following modifications: (i) we change the classifier size to $3336$ (number of classes), (ii) we add a dropout layer between the last fully-connected layer and the classifier, and (iii), we change the input size from $3$ channels (classical images classification) to $77$ channels according to the size of the input environmental tensors.
\end{itemize}

\noindent \textbf{Learning set up and parameters}: All our experiments were conducted using PyTorch deep learning framework\footnote{\url{https://pytorch.org/}} and were run on a single computing node equipped with 4 Nvidia GTX 1080 ti GPU. We used the Stochastic Gradient Descent optimization algorithm with a learning rate of 0.1 (divided by 10 at epoch 90, 130, 150 and 170), a momentum of 0.9, a mini-batch size of 128 and a dropout of 0.7. We perform a validation every 10 epochs and the final model is chosen as the one with the highest validation score.

\subsection{Spatial Models}
\label{spatial}
For this category of models, we rely solely on the spatial positions $p(x)$ to model the species distribution (\textit{i.e.} we do not use the environmental information at all). We did evaluate two different classifiers based on such spatial data:
\begin{enumerate}
    \item \textbf{Closest-location classifier}: For any occurrence $x_q$ in the test set and its associated spatial position $p(x_q)$, we return the labels of the species observed at the closest location $p_{NN}$ in $P_{train}$ (except $p(x_q)$ itself if $p(x_q) \in P_{train}$). The species are then ranked by their frequency of appearance at location $p_{NN}$. Note that $p(x_q)$ is excluded from the set of potential closest locations because of the construction protocol of the test. Indeed, as mentioned earlier, it was enforced that the occurrence of a given species in the test set does never occur at a location closer than 100 meters from the occurrences of the same species in the training set. As a consequence, if we took $p_{NN}=p(x_q)$, the right species would never belong to the predicted set of species.\\
    One of the problem of the above method is that it returns only a subset of species for a given query occurrence $x_q$ (\textit{i.e.} the ones located at $p_{NN}$. Returning a ranked list of all species in the training set would be more profitable with regard to the used evaluation metric (Mean Reciprocal Rank). Thus, to improve the overall performance, we extended the list of the closest species by the list of the most frequent species in the training set (up to reaching the authorized number of 100 predictions for each test item).
    
    \item \textbf{Random forest classifier}: Random forests are known to provide good performance on a large variety of tasks, including in ecology \cite{Cutler2007,GOBEYN2019179}, and are likely to outperform the naive closest-location based classifier described above. In particular we used the random forest algorithm implemented within the scikit-learn framework\footnote{\url{http://scikit-learn.org/stable/}}. We used only the spatial positions $p(x)$ as input variables and the species labels $y(x)$ as targets. For any occurrence $x_q$ in the test set, the random forest classifier predicts a ranked list of the most likely species according to $p(x_q)$. Concerning the hyper-parametrization of the method, we conducted a few validation tests on the training data and finally used $50$ trees of depth $8$ for the final runs submitted to the GeoLifeCLEF challenge. 
\end{enumerate}

\subsection{Co-occurrence Model}
\label{coocs}
Species co-occurrence is an important information in that it may capture inter-dependencies between species that are not explained by the observed environment. For instance, some species live in a community because they share preferences for a kind of environment that we do not observe (communities of weeds are often specialized to fine scale agronomic practices that are not reported in our environmental data), they use the available resources in a complementary way, or they favor one another by affecting the local environment (leguminous and graminaceous plants in permanent grasslands). On the opposite, some species are not likely to be observed jointly because they live in different environments, they compete for resources or negatively affect the environment for others (allelopathy, etc.).\\
\indent To capture this co-occurrence information, it is required to train a model aimed at predicting the likelihood of presence of a given species thanks to the knowledge of the presence of other species (without using the environmental information or the explicit spatial positions). Therefore, we did train a feed-forward neural network taking species \textit{abundance vectors} as input data and species labels as targets. The abundance vectors were built in a similar way than the closest-location classifier described in section \ref{spatial}. For any spatial position $q \in D$, we first aggregate all the occurrences located at the closest location $p_{NN}$ in $P_{train}$ (except $q$ itself). Then, we count the number of occurrences of each species in the aggregated set. More formally, we define the \textit{abundance vector} $\mathbf{z}(q) \in \mathbb{R}^N$ of any spatial position $q \in D$ as:
\begin{equation}
    \forall i,\forall x, z_i(q)= \sum_{p(x)=p_{NN}} \mathbb{1}(y(x)=i)
\end{equation}
where $\mathbb{1}()$ is an indicator function equals to 1 if the condition in parenthesis is true and $0$ otherwise and $z_i(q)$ is the component of $\mathbf{z}(q)$ corresponding to the abundance of the $i$-th species at position $q$.\\
\noindent \textbf{Architecture description}:
The neural network we used to predict the most likely species based on a given abundance vector is a simple Multi-Layered Perceptron (MLP) with one hidden layer of $256$ fully connected neurons. We used ReLU activation functions \cite{hinton2010relu} and Batch Normalization \cite{ioffe2015batch} for the hidden layer, and a softmax loss function as output of the network.\\
\noindent \textbf{Learning set up and parameters}:
This model was implemented and trained within PyTorch deep learning framework\footnote{\url{https://pytorch.org/}} and were run on a single computing node equipped with 4 Nvidia GTX 1080 ti GPU. We used the Stochastic Gradient Descent optimization algorithm with a learning rate of 0.001, a momentum of 0.9, a mini-batch size of 32. We perform a validation every epochs and the final model is chosen as the one with the highest validation score.

\subsection{Late Fusion of Previous Models}
\label{late_fusion}
We also produced an other model corresponding to a late fusion of the environmental CNN and the co-occurrence model. Indeed, the two base models being trained on different kinds of input data, we expect that their fusion may benefit from their complementarity. We process the late fusion by averaging the prediction probabilities of the two models and then we re-sort the predictions.

\section{New Model using Jointly Environment and Co-occurrences in an End-to-End Network}
\label{fusion model}

The previous neural networks allow to capture in one case information contained in plant co-occurrences and in the other case environmental information from patches. However, co-occurrences can also be seen as a complementary environmental information. Indeed, the species close to a plant are directly part of its environment. Thus, species interact with each other and with the physical environment. To capture the interdependencies between co-\-occur\-rences and the environment we have developed a model that uses both inputs: environmental tensors and vector of co-occurrences.\\

\noindent \textbf{Architecture description}: the environmental tensors are exactly the same as the CNN model and are described in section~\ref{data}.
The co-occurrences inputs remains the same as the co-occurrences model of the previous section, \textit{i.e.} the \textit{abundance vector} describe in section \ref{coocs}.
This model is a deep neural network that is partly convolutional. The architecture is describe in table~\ref{table:archi}. It is first composed of two separate branches. One branch of the network is identical to the environmental CNN with the inception v3 architecture without the dropout layer. The other branch of the network is a small neural network like the co-occurrence model consisting in two fully-connected layer of size 32 with a batch normalization after the first one. Instead of making two classifiers, the tensors from the last layers of the two branches are concatenated and, followed by a batch normalization and a dropout layer, and finally a single classifier on this final tensor. The role of this additional layer is to learn a common representation space that captures the potential interdependencies between the environment and the co-occurrences.\\
\\
\noindent \textbf{Learning set up and parameters}: This model is learned with the same setup, parameters and process than the environmental CNN (see section~\ref{CNN}) but with dropout of 0.8 instead of 0.7.

\begin{table}[ht]
\centering
\caption{Architecture of the fusion environmental and co-occurrences model}
\def\arraystretch{1.5}
\begin{tabular}{|c|}
  \hline
    \begin{tabular}{p{5.8cm}|p{5.8cm}}
        \large environmental tensors & \large co-occurrences vector\\
        \hdashline
        size: 64x64x77 & size: 3336\\
        \hline
        \large inception v3 & \large fully-connected (32) + batch normalization + fully-connected (32)\\
        \hdashline
        size: 2048 & size: 32\\
    \end{tabular}\\
  \hline
    \large concatenation + batch normalisation + dropout\\
    \hdashline
    size: 2080\\
\hline
    \large classifier\\
    \hdashline
    size: 3336\\
\hline
    \large prediction\\ 
\hline
\end{tabular}
\label{table:archi}
\end{table}

\section{Experiments and Results}
\label{expes}
\subsection{Evaluated Models Synthesis}
In summary, we evaluate the six following models:
\\
\textbf{Spa-CC}: the spatial closest-location classifier model (see section \ref{spatial}).\\
\textbf{Spa-RF}: the spatial random forest classifier (see section \ref{spatial}).\\
\textbf{Cooc-NN}: the co-occurrence model (see section \ref{coocs}).\\
\textbf{Env-CNN}: the environmental CNN (see section \ref{CNN}).\\
\textbf{Env-Cooc-LF}: late fusion of the probabilities given by the \textbf{Env-CNN} and the \textbf{Cooc-NN} models (see section \ref{late_fusion}).\\
\textbf{Env-Cooc-JNN}: joint neural network on environment and co-occurrences (see section \ref{fusion model}).\\

The 3 first ones (\textbf{Spa-CC}, \textbf{Spa-RF} and \textbf{Cooc-NN}) are exactly the same than submitted in our participation in the challenge (see \textbf{FLO\_1}, \textbf{FLO\_4} and \textbf{FLO\_2} in \cite{deneugeolifeclef2018}). \textbf{Env-CNN} is the new CNN architecture (see section \ref{CNN}) that was trained on the revised GeoLifeCLEF patches. \textbf{Env-Cooc-LF} is the same fusion model than \textbf{FLO\_5} in \cite{deneugeolifeclef2018} but with the new environmental CNN predictions. Finally, \textbf{Env-Cooc-JNN} is our new model no submitted in our participation.

\subsection{Model Selection and Validation Experiments}
\label{classif}
We conducted a first set of experiments to evaluate the performances of our models in the case of an independently and identically distributed validation set. Therefore, we extracted a part of the training set (10\% occurrences selected at random) and used it as a validation set. For neural networks we extracted an additional 10\% of remaining train occurrences to have a pre-validation set. We choose two cross-validation protocols :
\begin{itemize}
    \item For the three neural network models (\textit{i.e.} co-occurrences neural network, environmental CNN and environmental and co-occurrences fusion model) we choose to fix the split between training set, validation set and pre-validation set (Holdout cross-validation). As, the neural networks took around one day to be learned completely, it was not workable to repeat split and learning many times. Thus, we worked with a single validation set to calibrate all our neural networks models. By fixing the split we assume to introduce a bias, but this bias is then constant between the experiments which allows us to compare the performance obtained on a single learning.
    \item For the two spatial models, that require a lower computation time, we choose to not fix the train-validation split but to learn the model on twenty random train-validation splits (Monte Carlo cross-validation). The performance of a model is defined by the average performance of the model on the twenty different splits. Like this we don't introduce a bias as for the neural networks but we keep the possibility to compare two models. Note that for the random forest classifier of scikit-learn we need to have at least one occurrence of each species in the training set and one occurrence of each species in the test set. However, some species are present only once in the data, so we had to remove them for validation experiments of this model.
\end{itemize}

\subsection{Results}

\begin{table}
\centering
\caption{Results of our models on validation set and official test set.}
\begin{tabular}{|l|c|c|}
    \hline
    Model & validation MRR & official test set MRR\\ 
    \hline
    \textbf{Spa-CC} (run FLO\_1 of \cite{deneugeolifeclef2018})& 0.0640& 0.0199\\
    \textbf{Spa-RF} (run FLO\_4 of \cite{deneugeolifeclef2018})&  0.0781& 0.0329\\ 
    \textbf{Cooc-NN} (run FLO\_2 of \cite{deneugeolifeclef2018})&  0.0669& 0.0274\\ 
    \textbf{Env-CNN}& 0.0916&  0.0458\\
    \textbf{Env-Cooc-LF}& 0.0908& 0.0457\\
    \textbf{Env-Cooc-JNN}& 0.0927& 0.0479\\
    \hline
\end{tabular}
\label{results}
\end{table}

The performance of each model is given in tables \ref{results}. In validation, the best models are those based on environmental and co-occurrences data. The \textbf{Env-CNN} achieves a pretty good MRR of $0.0916$ knowing that the ideal MRR cannot exceed $0.409$ (due to the fact that several outputs exist for the same entry). On average, it returns the correct species in the first position with a success rate of 1/30 (0.0330) (knowing that there is $3336$ species in the training set). Concerning the two fusion models, if the late fusion did not result in a performance gain, the joint model (\textbf{Env-Cooc-JNN}) give a better score than the environmental model alone ($0.0927$ vs. $0.0916$) and is the best model evaluated here.\\
\indent Nevertheless, the other models achieve good results too, all are over $0.06$ of MRR and the random forest (\textbf{Spa-RF}) reaches almost $0.08$. They return the good species label between 1 time out of 40 and 1 time out of 30. These results show that some fairly simple models can capture a strong information.
For all models, the score on the official GeoLifeCLEF 2018 test set is much lower than the one obtained in validation. However, the order of performance of the models is maintained. The \textbf{Env-Cooc-JNN} model remains the best on the official test set, not far but significantly ahead of the \textbf{Env-CNN} model alone ($0.0479$ vs. $0.0458$).

These results allow us to draw the following conclusions:
\begin{itemize}
\item \textbf{official test set results vs. validation test set results}: overall, the MRR values achieved by our models on the blind test set of GeoLifeCLEF are much lower than the ones obtained within our validation experiments (see Table \ref{results}). We believe that this performance loss is mainly due to the construction of the blind test set, \textit{i.e.} to the fact that the occurrence of a given species in the test set does never occur at a location closer than 100 meters from the occurrences of the same species in the training set. This rule was not taken into account during our cross-validation experiments on the training set. An other point is that, for the official evaluation, the prediction size is limited at 100 ranked species for each test occurrences. We observe that this have also an impact on the MRR score.
\item \textbf{Supremacy of environmental models}: the results show that our models based on environmental data are the best performing ones. The environmental CNN model (\textbf{Env-CNN}) is ahead from co-occurrences and spatial models. The late fusion model on environment and co-occurrences (\textbf{Env-Cooc-LF}) obtained similar scores to the \textbf{Env-CNN} alone and the \textbf{Env-Cooc-JNN} outperform all other models on the two evaluation process. After the environmental models the spatial classifier based on random forest (\textbf{Spa-RF}) obtains a very fair performance considering that it only uses the spatial positions of the occurrences (which makes it very easy to implement in a real-world system). The co-occurrence model (\textbf{Cooc-NN}) obtains significantly lower performance, while the closest-location classifier, which uses only the nearest point species data, is the worst model (\textbf{Spa-CC}).
\item \textbf{The new environmental and co-occurrences model}: the joint environmental and co-occurrences model (\textbf{Env-Cooc-JNN}) is the best performing one. It allows a significant gain compared to the environmental model alone. This result indicates that there is a complementary information between co-occurrences and the environment. It also indicates that taking into account co-occurrences makes it possible to better characterize the ecological niche of the species compared to abiotic-only models.
\item \textbf{Score with new patches vs. old patches}:
as discussed in section \ref{data}, in this paper, we used a new set of environmental patches since a part of them was corrupted in the initial GeoLifeCLEF 2018 dataset. However, the results of this new study are still in accordance from the one reported in the context of the challenge (\cite{deneugeolifeclef2018}). In particular, the performance achieved by the environmental CNN on the new dataset confirms its superior predictive power over other proposed methods during the challenge. After patch correction, it remains the best model of the challenge, even better than the model learned on old environmental patches with a score of $0.0458$ against $0.0435$ (see \textbf{FLO\_3} in \cite{deneugeolifeclef2018}).
\item \textbf{Species community}: the co-occurrence model (\textbf{Cooc-NN}) seems to generalize better than the closest-location classifier (\textbf{Spa-CC}), though both methods used almost the same input information which is the species of the neighborhood. It is likely that the neural network detect the signature of a community from its input co-occurrences. For example, the network is able to predict a common Mediterranean species when it gets a rare Mediterranean species as entry. Indeed, the probability of observing this same rare species near its known observation is very small, but the closest location classifier would do the error.
\item \textbf{Non-performing late fusion}: the late fusion between the environmental model and the co-occurrence model (\textbf{Env-Cooc-LF}) did not result in a performance gain. However the \textbf{Env-Cooc-JNN} shows that there is some complementary information between co-occurrences and environment. This information seems to be therefore at the intersection of the environment and co-occurrences and requires joint learning to be effectively captured.
\end{itemize}

\section{Conclusion and Perspectives}
This paper compared four main types of models aimed at predicting species distribution: (i) a convolutional neural network trained on environmental variables extracted around the location of interest, (ii) a purely spatial model trained with a random forest, (iii) a co-occurrence based model aimed at predicting the likelihood of presence of a given species thanks to the knowledge of the presence of other species, and (iv), two fusions models between the environmental CNN and the co-occurrences model, one late fusion of predictions and one learned jointly on the to inputs. Our study shows that the convolutional neural network model maintains a high score with unbiased environmental patches. Indeed, it achieved the best performance over the others GeoLifeCLEF 2018 submitted models. However the main contribution of our study is the new joint model on environment and co-occurrences that achieve good results, significantly better than the environmental CNN. This shows that there is useful information in co-occurrences and that this information is at least partly complementary to environmental information. Few studies currently use this co-occurrences information. It would be interesting, in future work, to study more about how useful is the information in co-occurrences and how its complementarity with the environment can be explained.

\bibliographystyle{unsrt}
\bibliography{LifeCLEF}

\end{document}